\documentclass[10pt, a4paper]{article}

\usepackage{lrec-coling2024} 
\usepackage{amsmath}
\usepackage{amsthm}
\usepackage{booktabs}
\usepackage{algorithm}
\usepackage{algorithmic}
\usepackage{times}
\usepackage{amssymb}
\usepackage{graphicx}
\usepackage{caption}
\usepackage{booktabs}
\usepackage{hyperref}
\usepackage{tabularx}
\usepackage{adjustbox}
\usepackage{array}
\usepackage{multirow}
\usepackage{multicol}
\usepackage{subfigure}

\title{From Graph to Word Bag: Introducing Domain Knowledge to Confusing Charge Prediction}

\name{Ang Li\textsuperscript{1}, Qiangchao Chen\textsuperscript{2}, Yiquan Wu\textsuperscript{1}, Ming Cai\textsuperscript{1}, \\{\bf \large Xiang Zhou\textsuperscript{2\dag} \thanks{\textsuperscript{\dag} Corresponding author.}, Fei Wu\textsuperscript{1}, Kun Kuang\textsuperscript{1}}} 

\address{\textsuperscript{1}College of Computer Science and Technology, Zhejiang University \\
         \textsuperscript{2}GuangHua Law School, Zhejiang University \\
         \{liangrex, 22102078, wuyiquan, cm, 0020355, wufei, kunkuang\}@zju.edu.cn\\}

\abstract{
Confusing charge prediction is a challenging task in legal AI, which involves predicting confusing charges based on fact descriptions. While existing charge prediction methods have shown impressive performance, they face significant challenges when dealing with confusing charges, such as \emph{Snatch} and \emph{Robbery}. In the legal domain, constituent elements play a pivotal role in distinguishing confusing charges. Constituent elements are fundamental behaviors underlying criminal punishment and have subtle distinctions among charges. In this paper, we introduce a novel \textbf{F}rom \textbf{G}raph to \textbf{W}ord \textbf{B}ag (FWGB) approach, which introduces domain knowledge regarding constituent elements to guide the model in making judgments on confusing charges, much like a judge's reasoning process. Specifically, we first construct a legal knowledge graph containing constituent elements to help select keywords for each charge, forming a word bag. Subsequently, to guide the model's attention towards the differentiating information for each charge within the context, we expand the attention mechanism and introduce a new loss function with attention supervision through words in the word bag. We construct the confusing charges dataset from real-world judicial documents. Experiments demonstrate the effectiveness of our method, especially in maintaining exceptional performance in imbalanced label distributions.
 \\ \newline  }
\begin{document}

\maketitleabstract

\section{Introduction}

In recent years, artificial intelligence has been applied in the legal domain. Legal artificial intelligence (LegalAI) focuses on applying artificial intelligence methods to benefit legal tasks \cite{zhong-etal-2020-nlp}. These advancements have led to increased research on charge prediction \citep{ bib1, luo-etal-2017-learning, attributes2018,ye-etal-2018-interpretable}. In all these studies, researchers approach the charge prediction task as a classification problem, utilizing classification models. Substantial progress has been made in these areas over time.

In charge prediction, many methods have been extensively proposed and they have commendable predictive performance. However, this task still faces challenges when dealing with confusing charges in real legal scenarios. Most existing methods for crime prediction primarily focus on the legal system structure \cite{bib14} or on macro-level semantic knowledge \cite{attributes2018}. These methods utilize legal knowledge to assist the model, but they are not sufficient to enable the model to master the ability to distinguish between confusing charges.

\begin{figure*}[t]
    \centering
    \includegraphics[width=1.0\textwidth]{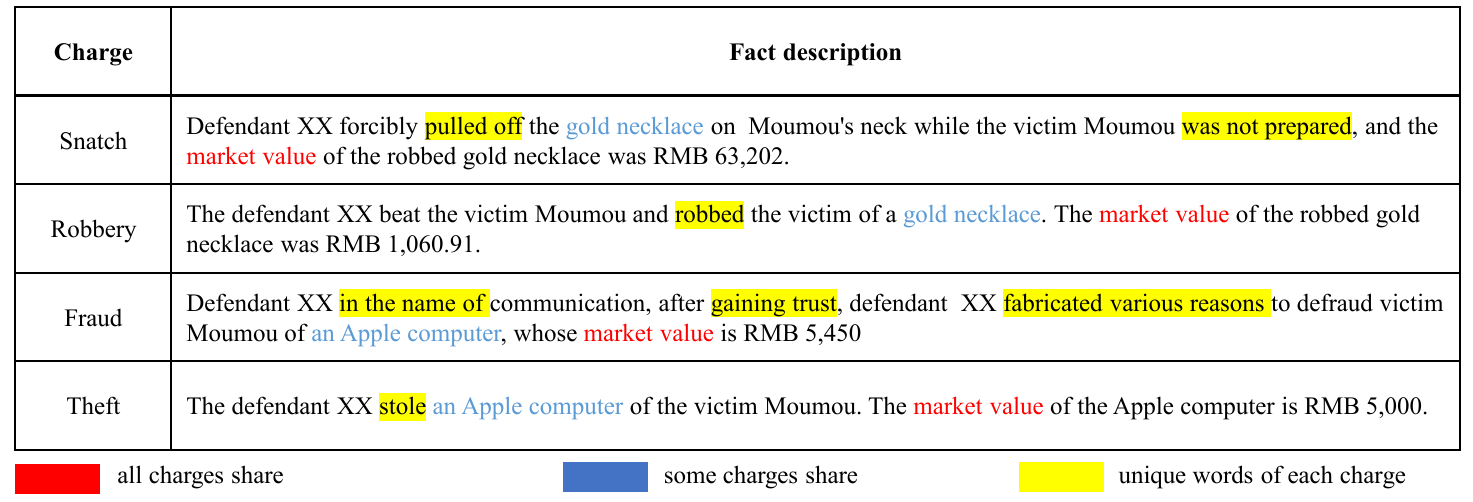}
    \vspace{-15pt}
    \caption{Confusing charges in real legal cases. Red words indicate the words all charges share, blue words indicate the words some charges share, and the yellow highlighted words indicate the unique words of each charge.}  
    \vspace{-10pt}
    \label{example}
\end{figure*}

In this work, we focus on the task of confusing charge prediction. which is a subset of charge prediction, specifically addressing cases where the performance is poor. Fig. \ref{example} illustrates a cluster of confusing charges in real cases, and highlights shared words and unique words for each charge. From a legal perspective, the key to distinguishing these charges can be reflected through unique words: \emph{Theft} is characterized by non-violence, \emph{Snatch} involves violence against property, \emph{Robbery} entails violence against individuals, and \emph{Fraud} includes descriptions related to trust. Therefore, the challenge in this task is: How to make the model focus on and understand the critical information that distinguishes confusing charges?

It's worth noting that the constituent elements play a pivotal role in charge prediction. Constituent elements refer to the types of behavior or crimes that serve as the basis for criminal punishment according to abstract provisions of criminal law. To address the aforementioned challenge, we propose a novel \textbf{F}rom \textbf{G}raph to \textbf{W}ord \textbf{B}ag (FWGB) approach to leverage these constituent elements. Specifically, we construct a knowledge graph that encompasses distinguishing constituent elements. Through a graph-based keyword selection method, we automatically extract words highly relevant to the constituent elements in the knowledge graph, thus forming a word bag. Subsequently, to make the model pay more attention to the distinguishing information for each charge within the context, we propose a multi-attention supervision method. Specifically, we expand the attention mechanism and introduce a new loss function with attention supervision through words in the word bag.

To verify the effectiveness of our method, we construct the confusing charge dataset by selecting easily confusing charges based on real-world data. The comparison with numerous powerful baselines demonstrates the effectiveness of our method, as it outperforms many strong baselines. Ablation experiments show that using a legal knowledge graph with constituent elements can enable the model to learn more distinguishing knowledge about the confusion charges. The multi-attention supervision can help the model focus on distinguishing information in the context. It is worth noting that we are the first to use a legal knowledge graph with constituent elements to assist in charge prediction.


In summary, we make the following contributions:
\begin{itemize}
\item[$\bullet$] We investigate the task of confusing charge prediction by taking the domain knowledge into consideration. 

\item[$\bullet$] We propose a novel \textbf{F}rom \textbf{G}raph to \textbf{W}ord \textbf{B}ag (FWGB) approach. Specifically, we construct an expert knowledge graph with constituent elements and then form the word bag, combining multi-attention supervision to guide the model in distinguishing between confusing charges.
\item[$\bullet$] We construct the confusing charge dataset from real-world data. Our experiments evaluate the effectiveness of our proposed method. We make the code and dataset publicly available \footnote{\href{https://github.com/LIANG-star177/FWGB}{https://github.com/LIANG-star177/FWGB}} for reproducibility.
\end{itemize}

\vspace{-5pt}
\section{Related Work}
  \subsection{AI and Law}
AI and law is an emerging interdisciplinary field of law and computer science. Currently, several scholars focus on the regulation of AI by law \citep{bib6}, while others choose to study the application of AI techniques in the field of law. The most studied tasks of the applications are legal judgment prediction\citep{bib7, bib33,LiuWZS0WK23,wu-etal-2023-precedent}, legal question answering\citep{bib8}, legal case retrieval\citep{bib9}, legal information extraction\citep{bib32} and legal summarization\citep{bib10}. Legal judgment prediction aims to provide legal consequences, including the charges, prison terms, and so on, for professionals to lighten their workload or for laymen to learn about the case they are concerned about. Our work focuses on confusing charge prediction, which is one of the aspects of legal judgment prediction.

  \subsection{Charge Prediction}
  \vspace{-5pt}
Charge prediction is a subtask of legal judgment prediction that takes fact descriptions as the input of the model and charges as the output of the model. Early work focused on predicting charges through artificial intelligence analysis of charge features \cite{bib17} or through manually designed methods \cite{bib18}. However, due to the large amount of feature engineering of these methods, Some researchers proposed leveraging the legal system's structure or incorporating emerging technologies like graph neural networks to enhance task performance \cite{bib14,bib29,bib4,bib19,yue2021neurjudge,dong2021legal}. Simultaneously, with pretrained models similar to BERT \cite{devlin2019bert} achieving outstanding performance in many classification tasks, some models specifically pretrained for the legal domain have also been introduced \cite{XIAO202179,bib28}. Though this task has been explored for a long time, confusing charge prediction still needs to be improved. \citet{an2022charge} define confusing charges: If two charges differ in only one constitutive element, they are considered confusing charges to each other. In this work, We enhance confusing charge prediction by introducing a word bag formed by the knowledge graph with constituent elements, coupled with a multi-attention mechanism for model supervision. Moreover, Our model gets interpretability from inherent legal knowledge, allowing it to make predictions in a lawyer's way.

\vspace{-10pt}
\begin{figure*}[t]
    \centering
    \includegraphics[width=1\textwidth]{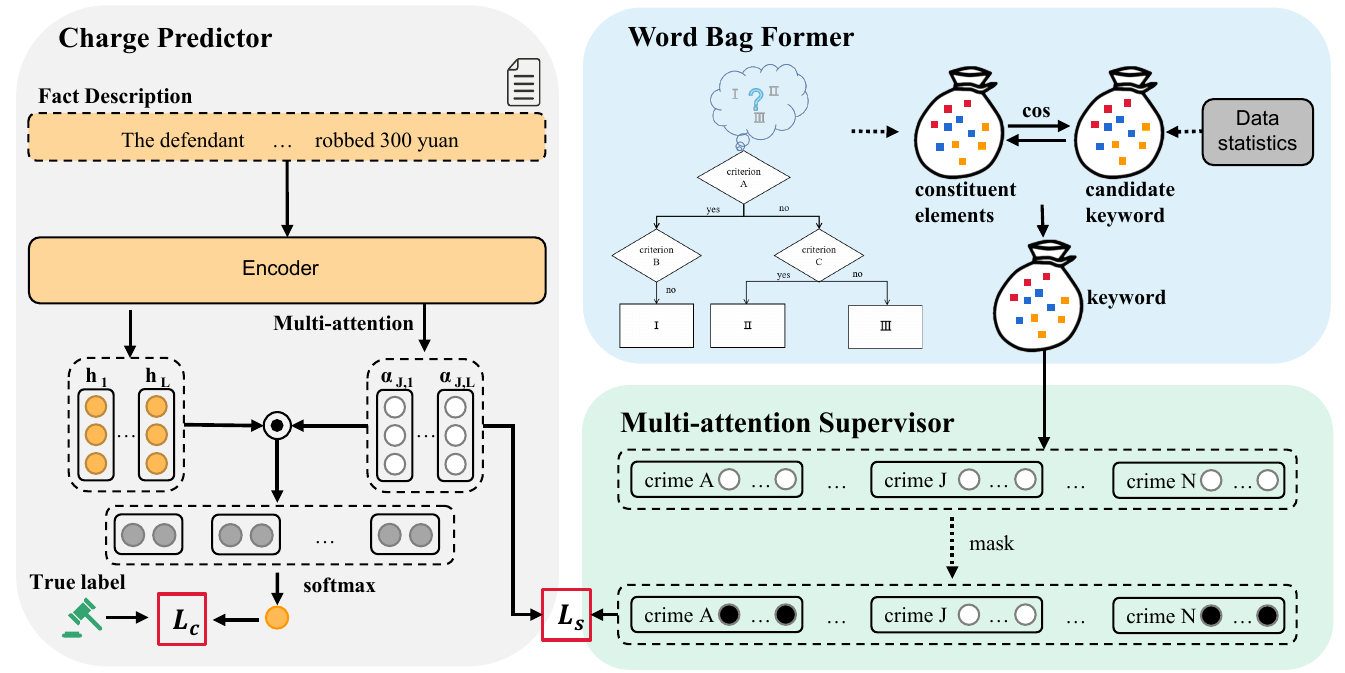}
    \caption{Structure of Our Model. The charge predictor uses LSTM to encode fact descriptions, employs a multi-attention mechanism for label-independent attention scores, and derives probability distributions for each label. The word bag former transforms expert knowledge graphs into prerequisites, selecting genuine keywords from statistical data to create a word bag. The Multi-attention supervisor assumes high attention values for label-related keywords, masking out irrelevant ones to guide the attention mechanism. Here, $L_c$ is the loss of classification, and $L_s$ is the loss associated with attention supervision.}
    \label{fig:Structure}
\end{figure*}
\section{Method}
As illustrated in Fig. \ref{fig:Structure}, our method FGWB comprises three key components: Charge predictor, Word bag former, and Multi-attention supervisor.
  \subsection{Word Bag former}
    \subsubsection{Using Constituent Elements to Form Expert Knowledge Graphs}
One legal domain knowledge frequently used in real legal scenarios is constituent elements. Constituent elements refer to the preconditions that a certain behavior should have to be evaluated by law. In brief, when a person's behavior in life meets the constituent elements of the law, the person may assume legal responsibility. For example, constituent elements of fraud are: (1) The suspect commits a fraudulent act; (2) The victim falls into a wrong understanding, and so on.
However, expert domain knowledge such as constituent elements is too obscure for laymen. Inspired by \citet{bib31}, we construct the knowledge graph to use constituent elements easily, as shown in Fig. \ref{fig:gragh}. Specifically, we use a set of charges as nodes and employ the constituent elements as the connections between nodes. To allow people to clearly understand significant features, we downplay some elements in the graph. For instance, the constituent elements of fraud contain five while theft contains different two. But legal experts can distinguish the two crimes only relying on the differences in the disposal act, which is indicted through ``fraudulent act" and ``stealing act".

    \subsubsection{Graph-based Keyword Selection}
To make the model focus on differentiating information, a crucial prerequisite is to identify distinguishing keywords in fact descriptions, and thus address the issue of charge confusion. A simple method to find keywords is based on data statistics. However, it has two issues: (1) Different criminal charges may share a set of common keywords, and focusing on these words does not resolve the confusion problem. (2) Some words may not have a genuine relationship with the criminal charge, leading to misconceptions.

\begin{figure}[t]
    \centering
    \includegraphics[width=0.5\textwidth]{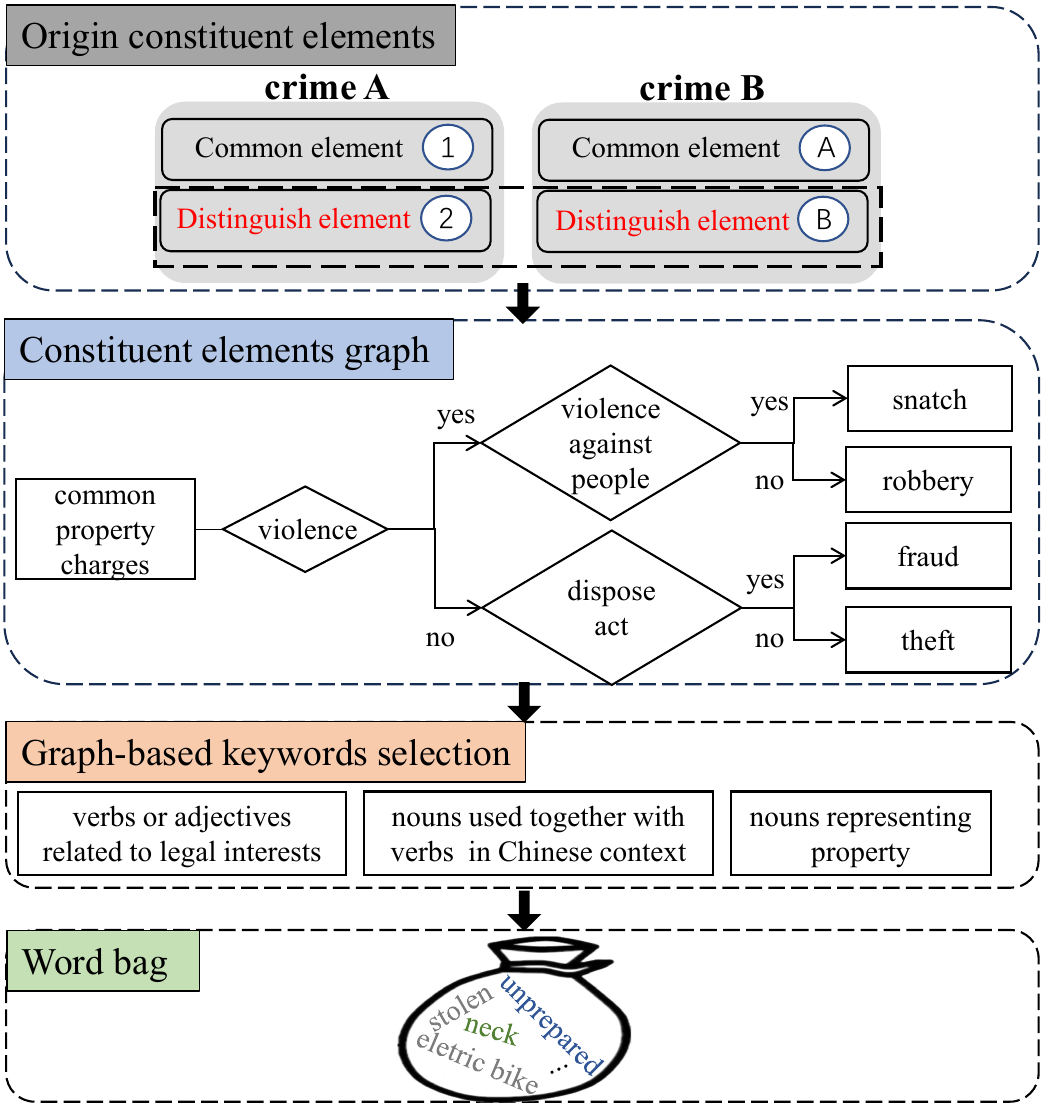}
    \caption{Construction and utilization of expert knowledge graphs.}
    \label{fig:gragh}
\end{figure}

We notice that all the constituent elements along the path from the starting node to the leaf nodes in the knowledge graph can comprehensively describe a criminal charge. For example, in Fig. \ref{fig:gragh}, the crime of \emph{Snatch} in the graph includes the \emph{violence} and \emph{violence against people} constituent elements. Therefore, to find actual keywords, we use the constituent elements in the knowledge graph to select keywords obtained from data statistics. 

For a given criminal charge $i$, we first use the data statistics method to filter out a candidate keyword set $C_i^{'}$. We then obtain the constituent elements set $R_i$ of criminal charge $i$ from the knowledge graph. We feed both the candidate keywords and the constituent elements into a legal pretrained model $f_\theta$ to obtain their corresponding vector representations. For each word in $C_i^{'}$, we calculate its cosine similarity with each word in $R_i$. If the average similarity between the word and the constituent element set exceeds the threshold $\eta$, we consider that word as a keyword of true keyword set $C_i$. 
\begin{equation}
   \frac{{\textstyle \sum_{w\epsilon C_i^{'}, r\epsilon R_i}} sim(f_\theta(w),f_\theta(r))}{|R_i|} >\eta \to w\epsilon C_i
\end{equation}
Subsequently, we filter out words that genuinely characterize the constituent elements, resulting in the final word bag $B=\{C_1,...,C_N\}$, where $N$ is the number of charge labels. It's worth noting that the automatic formation of the word bag is independent of the model and only needs to be executed once.

\subsection{Charge Predictor}
\subsubsection{Encoder}
To assess the generality of our approach, we implemented the attention mechanism with supervision on two different encoders. Specifically, we employed LSTM \cite{liu2019bidirectional} trained from scratch and Electra \cite{bib28}, which was pretrained on legal data. When a L-length word sequence x=\{$x_{1}, x_2, ...,x_{L}$\} is put into the encoder, every word $x_{i} \in $ x is converted to its hidden state $h_{i}$ according to the following formulas:
\begin{equation}
\{h_{1}, h_2, ...,h_{L}\} = Encoder\{x_{1}, x_2, ...,x_{L}\}
\end{equation}

\subsubsection{Multi-attention Mechanism}
Because the keywords in the word bag contain the key information that determines the charge prediction, we naturally think of using the attention mechanism to let the model pay attention to the key information. To get independent attention, we introduce a multi-attention mechanism on top of the basic attention mechanism. Here, we introduce the acquisition of the keyword attention matrix, and the supervision methods are implemented through training loss functions in Sec. \ref{sec:training}. 

\paragraph{\textbf{Attention}.} Firstly, we input the hidden state into the single-layer neural network to obtain a vector, then multiply the transpose of the vector and the context vector, and obtain the $i$-th token's attention weight after softmax normalization. The formula is shown as follows:
\begin{equation}
a_{i}=\frac{exp(tanh(Wh_i+b)^{T}u)}{\sum^{L}_i exp(tanh(Wh_i+b)^{T}u)}
\end{equation}

where $h_i$ is the hidden state, $W$, $b$ are trainable parameters; $u$ is the context vector. $a=\{a_1,..., a_i,..., a_L\}$ is attention sequence. 

\paragraph{\textbf{Multi-attention}.} Traditional attention pays attention to every word in the word bag while in legal practice, not all words relate to a certain charge. Thus, we propose the idea to use multi-attention to meet the needs of legal practice.

Given the Word bag $B=\{C_1,...,C_N\}$ contains a total of $R$ words, among which there are $N$ charges. In order to independently compute the attention for each charge, we expand the context vector $u$ to $N$ dimensions, corresponding to the number of charges. For the $n$-th charge, $n \in [1, N]$, the corresponding attention weights are calculated as follows:
\begin{equation}
\alpha_{i,n}=\frac{exp(tanh(Wh_i+b)^{T}u_{n})}{\sum^{L}_i exp(tanh(Wh_i+b)^{T}u_{n})}
\end{equation}

where $u_{n}$ is the context vector corresponding to the $n$-th charge, $a=\{a_{1,1},..., a_{i,n},..., a_{L,N}\}\in \mathbb{R}^{L \times N}$ is attention matrix. 

\subsubsection{Predictor}
The predictor calculates the weighted average hidden produced by the hidden state from the encoder and attention weights, for attention:
\begin{equation}
s=\textstyle \sum^{L}_{i}\alpha_{i}h_{i},
\end{equation}
for multi-attention:
\begin{equation}
s=\textstyle \sum^{L}_{i} \sum^{N}_{n} a_{i,n}h_{i}.
\end{equation}

The charge predictor predicts the distribution of $y$ overall charges through one fully connected layer and a softmax function:
\begin{equation}
y=softmax(Ws+b)
\label{eq:y}
\end{equation}

\subsection{Training}
\label{sec:training}
\subsubsection{Multi-attention Supervisor}
 To guide the model's focus toward crucial information that can distinguish between confusing charges, we design an attention supervision loss in the training stage. Similarly, we first introduce the basic attention supervision and then proceed to introduce our multi-attention supervision.
\paragraph{\textbf{Attention Supervision Loss}.}
First, we prepare a target attention sequence $\hat{a}=[\hat{a}_{1}, ..., \hat{a}_{i}, ..., \hat{a}_{L}]$, L is the length of the input. Specifically, if a word from the input is present in the word bag, we set its target attention value to 1; otherwise, it is set to 0. Then, to guide the model's focus on keywords, we calculate the loss between the target attention sequence and the calculated attention values $a=[a_{1}, ..., a_{i}, ..., a_{L}]$ as follows:
\begin{equation}
\mathcal{L}_{s}=-\sum^{L}_{i=1}[\hat{a}_{i}log(a_{i})+(1-\hat{a}_{i})log(1-a_{i})]
\label{eq:loss_key}
\end{equation}

\paragraph{Multi-attention Supervision Loss.}
To enable the model to independently direct its attention to distinct key information when dealing with various criminal charges, we supervise the multi-attention values by a loss function. Here, we establish a target attention as an $N*L$ matrix, where $\hat{a}_{i,n}$ represents the attention value of the $i$-th word corresponding to the $n$-th criminal charge label. Similarly, the calculated attention is also an $N*L$ matrix. Although the dimensions of these attention matrices are $N*L
$, we mask the $(N-1)*L$ attention weights that do not belong to the current charge label according to the true label:
\begin{equation}
a_{i}=\mathrm{MASK}(n)(a_{1,1}, ..., a_{i,n}, ..., a_{L,N})=a_{i,n}
\end{equation}
Where $\mathrm{MASK}(n)$ indicates that the current sample is labeled as $n$, only the attention value $a_{i,n}$ is retained. That's to say, after the masking operation, the matrix we need to supervise remains a sequence, and the dimension is the same as that of the traditional attention supervision, so the loss function $\mathcal{L}_s$ is also the same as Eq. \ref{eq:loss_key}.

\subsubsection{Total Loss}
The total loss of our model contains two parts: $\mathcal{L}_{s}$ and $\mathcal{L}_{c}$. $\mathcal{L}_{s}$ is the loss function to supervise the attention of keywords, while $\mathcal{L}_{c}$ is the loss function to minimize the cross entropy between the ground-truth $y$ and predicted charge label $\hat{y}$ as follow:
\begin{equation}
\mathcal{L}_{c}=-yln\hat{y}
\end{equation}
\par The total loss is the sum of $\mathcal{L}_{s}$ and $\mathcal{L}_{c}$, $\lambda$ is the adjustment coefficient for attention supervision:
\begin{equation}
\mathcal{L}=\mathcal{L}_{c}+\lambda\cdot\mathcal{L}_{s}
\end{equation}

\begin{table}[t]
\small
\centering
\begin{tabular}{lcc}
\toprule
Type & Property Set & Drug Set\\
\midrule
\# Train set & 34529 & 11391\\
\# Valid set & 3836 & 1266\\
\# Test set & 4000 & 2000\\
Avg. \# Tokens in Fact & 339 & 347\\
\bottomrule
\end{tabular}
\caption{Data Set Collection}
\label{tab:number}%
\end{table}

\begin{table}[t]
\small
\centering
\begin{tabular}{lc|lc}
\toprule
Property Set & Number & Drug Set & Number\\
\midrule
Theft & 28535 & DS & 5345\\
Fraud & 8426 & PVFDU & 4789\\
Robbery & 1048 & IPOD & 1668\\
Snatch & 356 & DT & 855\\
\bottomrule
\end{tabular}
\caption{Label Distribution}
\label{tab:label}%
\end{table}

\begin{table*}[t]
\small
\centering
\begin{tabular}{l|ccccc|ccccc}
\toprule
\multicolumn{1}{l|}{\multirow{2}{*}{Method}} & \multicolumn{5}{c|}{Property Set} &\multicolumn{5}{c}{Drug Set}\\

\multicolumn{1}{c|}{\multirow{2}{*}{}} & Ma-P& Ma-R&Ma-F & Acc &Ma-F* & Ma-P& Ma-R&Ma-F & Acc &Ma-F* \\
\midrule
TextRNN  & 0.789 & 0.784  &0.769 
& 0.783  &0.769 & 0.913 & 0.910 & 0.909  &0.909& 0.895  \\
LSTM & 0.857& 0.815&0.808& 0.814&0.798& 0.914& 0.901& 0.909&0.924& 0.914\\
TextCNN & 0.846 & 0.799  &0.776 
& 0.800  &0.743 & 0.903 &	0.880 &	0.878  &0.884&	0.908 \\
DPCNN & 0.882 & 0.864  &0.856 & 0.865  &0.896 &	0.931&0.934&0.927&0.937&0.947\\
C3VG  & 0.882 & 0.862  &0.860 & 0.868  &0.985 & 0.920 & 0.920 & 0.920  &0.914& 0.939\\
Electra & 0.903 & 0.889  &0.881 & 0.888  &0.892 & 0.936 & 0.930 & 0.928  &0.928& 0.943 \\
Topjudge & 0.891 & 0.877  &0.874 & 0.878  &0.886 & 0.926& 0.923& 0.922&0.917& 0.940\\
LADAN  & 0.905 & 0.893&0.892& 0.895&0.908 & 0.933& 0.930& 0.929&0.940& 0.966 \\
NeurJudge  & \underline{0.907} & \underline{0.897}  &\underline{0.902} & \underline{0.905}  &\underline{0.917} & \underline{0.939}& 0.940& 0.935&0.950 
& 0.968\\
R-Former & 0.905 & 0.895  &0.894 
& 0.901  &0.918 & 0.931& \underline{0.948}& \underline{0.941}&\underline{0.951}& \underline{0.970}\\
\midrule
FGWB (LSTM) & 0.888 & 0.880  &0.876 
& 0.880  &0.880 & 0.934& 0.933& 0.930&0.929& 0.942\\
FGWB (Electra) & \textbf{0.923} & \textbf{0.925}  &\textbf{0.924} & \textbf{0.925}  &\textbf{0.928} & \textbf{0.957} & \textbf{0.955} & \textbf{0.956}  &\textbf{0.955} & \textbf{0.979} \\
\bottomrule
\end{tabular}
\caption{Experiment results for property charges and drug charges, the best is \textbf{bolded} and the second best is \underline{underlined}.}
\label{results}
\end{table*}

\section{Experiment}
  \subsection{Dataset Description}
 We collect our data from 12309 China Procuratorate Website \footnote{https://www.12309.gov.cn}. Considering that our study focuses on confusing charges in real law scenes, we choose common property charges, including \emph{Theft}, \emph{Fraud}, \emph{Snatch}, and \emph{Robbery}, which are easily confused. After disposing of accusations with garbled or incomplete contents and multiple charges or multiple defendants, we finally got 38365 cases to form a data set. Further, we observe the distribution of charges is imbalanced. The number of \emph{Theft} is 80 times that of \emph{Snatch}, which indicates the imbalance between charges. To fairly assess the model's performance, we additionally construct a balanced test set from CAIL2018\footnote{http://cail.cipsc.org.cn/index.html} \citeplanguageresource{xiao2018cail2018}. Specifically, we set the number of cases for each charge in the test set to 1000. Then for the rest of the cases, we randomly divided them into a training set and validation set at the ratio of 9:1. To assess the model's ability to handle imbalanced distributions, we used the validation set as an imbalanced test set for comparison.

 To verify the generality, we apply the same processing steps to another cluster of common drug charges to get the second data set, containing \emph{Drugs Selling (DS)}, \emph{Providing Venues For Drug Users (PVFDU)}, \emph{Illegal Possession Of Drugs (IPOD)} and \emph{Drugs Transportation (DT)} charges. we set the number of cases for each charge in the test set to 500. The details of the dataset are shown in Tab. \ref{tab:number} and Tab. \ref{tab:label}.


  \subsection{Baselines}
To evaluate the performance and interpretability of our model, we implemented several baselines to compare these two aspects. \textbf{TextRNN} \citep{graves2013generating} is a traditional recurrent neural network model for text classification. \textbf{LSTM} \cite{zhou2015c} incorporates both forward and backward information flow through LSTM units to capture contextual information effectively. \textbf{TextCNN} \citep{lai2015recurrent} is a traditional convolutional neural network model for text classification. \textbf{DPCNN} \citep{bib30} is a low-complexity word-level deep convolutional neural network architecture for text categorization. \textbf{C3VG} \citep{bib29} is a model following a two-stage architecture which is from extraction to generation. \textbf{Electra} \citep{bib28} is a pretrained model that has been adjusted to Chinese. \textbf{Topjudge} \citep{bib14} is a model that incorporates multiple subtasks and DAG dependencies into judgment prediction. \textbf{LADAN} \citep{bib4} is a model that attentively extracts features from law cases' fact descriptions to distinguish confusing law articles. \textbf{NeurJudge} \citep{yue2021neurjudge} splits the fact description into two parts and encodes them separately. \textbf{R-Former} \citep{dong2021legal} formalizes LJP as a node classification problem.

To further validate our proposed FWGB, we conducted three sets of ablation experiments for each of the two encoder methods: \textbf{w/o SV} means not using attention supervision but retaining the multi-attention configuration. \textbf{w/o Multi-Attn} uses traditional attention mechanisms and applies attention supervision. \textbf{w/o KG} employs the multi-attention supervision mechanism but does not use the knowledge graph to filter the word bag, instead using a word bag composed of high-frequency words.

\subsection{Experimental Settings}
Our experiment is carried out on two V100 GPUs, and all the baseline models adopt the settings in their original papers. 
For the models without pretrained models, we adopt Gensim \citep{rehurek_lrec} on the training corpus to initialize the word embeddings, which are in the dimension of 300. 
For samples with long input, we truncate them to 512 tokens. We set the coefficient $\lambda$ to the best-performing 0.7 and explore the impact of different values of $\lambda$ on performance.

To evaluate the performance of the prediction, we calculate the Macro precision (Ma-P), Macro recall(Ma-R), and Macro F1 score (Ma-F) and accuracy (Acc). Ma-F* is used to represent the macro-F1 score tested on the imbalanced test set.

\begin{table*}[t]
\small
\centering
\begin{tabular}{l|ccccc|ccccc}
\toprule
\multicolumn{1}{l|}{\multirow{2}{*}{Method}} & \multicolumn{5}{c|}{property charges} &\multicolumn{5}{c}{drug charges}\\

\multicolumn{1}{c|}{\multirow{2}{*}{}} & Ma-P& Ma-R&Ma-F & Acc &Ma-F* & Ma-P& Ma-R&Ma-F & Acc &Ma-F* \\
\midrule
FGWB (LSTM)  & \textbf{0.888} & \textbf{0.880 } &\textbf{0.876} 
& \textbf{0.880}  &\textbf{0.880} & \textbf{0.934}& \textbf{0.933}& \textbf{0.930}&\textbf{0.929}& 0.942\\
w/o SV  & 0.862 & 0.839  &0.857
& 0.848  &0.872 
& 0.914 & 0.897& 0.911& 0.904& 0.924\\
w/o Multi-Attn  & 0.866 & 0.842  &0.861 
& 0.854  &0.862 & 0.918 & 0.908& 0.917& 0.921& 0.931 \\
w/o KG & 0.874& 0.869&0.863& 0.865&0.872&  0.930 & 0.923& 0.921& 0.919& \textbf{0.943}\\
\midrule
FGWB (Electra)  & \textbf{0.923} & \textbf{0.925}  &\textbf{0.924} & \textbf{0.925}  &\textbf{0.938} & \textbf{0.957} & \textbf{0.955} & \textbf{0.956}  &0.955 & \textbf{0.979}\\
w/o SV  & 0.901 & 0.892  &0.907
& 0.897  &0.905 
& 0.942& 0.904& 0.920& 0.919& 0.957\\
w/o Multi-Attn  & 0.907 & 0.897  &0.917 
& 0.902  &0.911 
& 0.948& 0.927& 0.934& \textbf{0.959}& 0.965 \\
w/o KG & 0.912 & 0.902  &0.910 
& 0.895  &0.919
& 0.950& 0.948& 0.935& 0.948& 0.977 \\
\bottomrule
\end{tabular}
\caption{Ablation Experiment Results}
\label{ablation}
\end{table*}

\begin{table}[t]
\small
\begin{tabular}{llll}
\begin{tabular}{l|cccc}
\hline
NeurJudge & Robbery & Snatch & Theft & Fraud \\
\hline
Robbery & 709 & 7 & 0 & 2 \\
Snatch & 48 & 836 & 1 & 2 \\
Theft & 108 & 41 & 989 & 6 \\
Fraud & 2 & 10 & 10 & 992 \\
\hline
\end{tabular}
\\
\multicolumn{1}{c}{} \\
\begin{tabular}{l|cccc}
\hline
Electra & Robbery & Snatch & Theft & Fraud\\
\hline
Robbery & 888 & 244 & 1 & 0 \\
Snatch & 4 & 355 & 2 & 0 \\
Theft & 101 & 376 & 982 & 10 \\
Fraud & 7 & 25 & 15 & 990 \\
\hline
\end{tabular}
\\
\multicolumn{1}{c}{} \\
\begin{tabular}{l|cccc}
\hline
FGWB (SV) & Robbery & Snatch & Theft & Fraud \\
\hline
Robbery & 872 & 71 & 0 & 0 \\
Snatch & 68 & 862 & 2 & 0 \\
Theft & 35 & 34 & \textbf{997} & 6 \\
Fraud & 25 & 33 & 1 & 994 \\
\hline
\end{tabular}
\\
\multicolumn{1}{c}{} \\
\begin{tabular}{l|cccc}
\hline
FGWB (MSV) & Robbery & Snatch & Theft & Fraud \\
\hline
Robbery & \textbf{902} & 29 & 0 & 0 \\
Snatch & 181 & \textbf{942} & 1 & 0 \\
Theft & 45 & 98 & 996 & 3 \\
Fraud & 5 & 37 & 3 & \textbf{995} \\
\hline
\end{tabular}
\end{tabular}
\caption{Confusing Matrices for Different Models. ``SV" stands for using attention supervision, while ``MSV" stands for using multi-attention supervision. They are both implemented on Electra.}
\label{confusion_matrices}
\end{table}

\subsection{Experiment Results}

\paragraph{Result of Charge Prediction:} From Tab. \ref{results} of Property Set, we observe that:
(1) FGWB (Electra) model significantly and consistently outperforms all the baselines. The result proves that FWGB effectively draws the model's attention to specific parts of fact descriptions that can help to make the right predictions.
(2) Compared with baselines without attention supervision mechanism, FGWB (Electra) gets 1.6\% more scores on Ma-P, 0.020 more scores on recall, 2.8\% more scores on Ma-R, 2.2\% more scores on Ma-F and 2.0\% more scores on Acc than the best-performed baseline (NeurJudge). 
(3) Compared with FGWB (LSTM), FGWB (Electra) gets 4.8\% more scores on Ma-F. This indicates that pretrained models acquire knowledge that is beneficial for helping the model understand input information. 

Comparing the result of the imbalanced test set (valid set) and the balanced test set, we conclude that: 
(1) Ma-F* values are higher than those of Ma-F, which indicates that an imbalanced data set is a huge challenge for charge prediction. 
(2) By comparing the difference between Ma-F* and Ma-F for each method, we find that the FGWB (Electra) is the most suitable model for an imbalanced data set with the fact that it gets the least decrease of 0.4\%.

From Tab. \ref{results} of Property Set, we get the similar observations:
(1) Ma-P, Ma-R, Ma-F and Acc value for FGWB (Electra) win. the best-performed baseline R-Former, which is a corroboration for the validity of our method.
(2) Our FWGB method exhibits significant improvements in both implementation approaches (LSTM and Electra). In summary, our approach performs well on the dataset containing drug-related confusing charges, highlighting its generalizability and applicability to various situations.

\paragraph{Result of Confusing Matrices:} From the confusing matrices of different models shown in Tab. \ref{confusion_matrices}, We can conclude that:
(1) Compared FGWB with the model without attention supervision mechanism(LSTM, NeurJudge), methods that use attention supervision are generally effective at improving the model's performance for confusing charges. On low-frequency charges, FGWB (MSV) gets 106 more right predictions on \emph{Snatch}. This indicates that our approach can better handle label imbalance compared to other methods.
(2) The confusing matrix of FGWB (MSV) outperforms that of FGWB (SV) on all charges, which shows the attention trained under the supervision of legal knowledge is better than the traditional attention.

\begin{figure*}[t]
\centering
\includegraphics[width=1.0\textwidth]{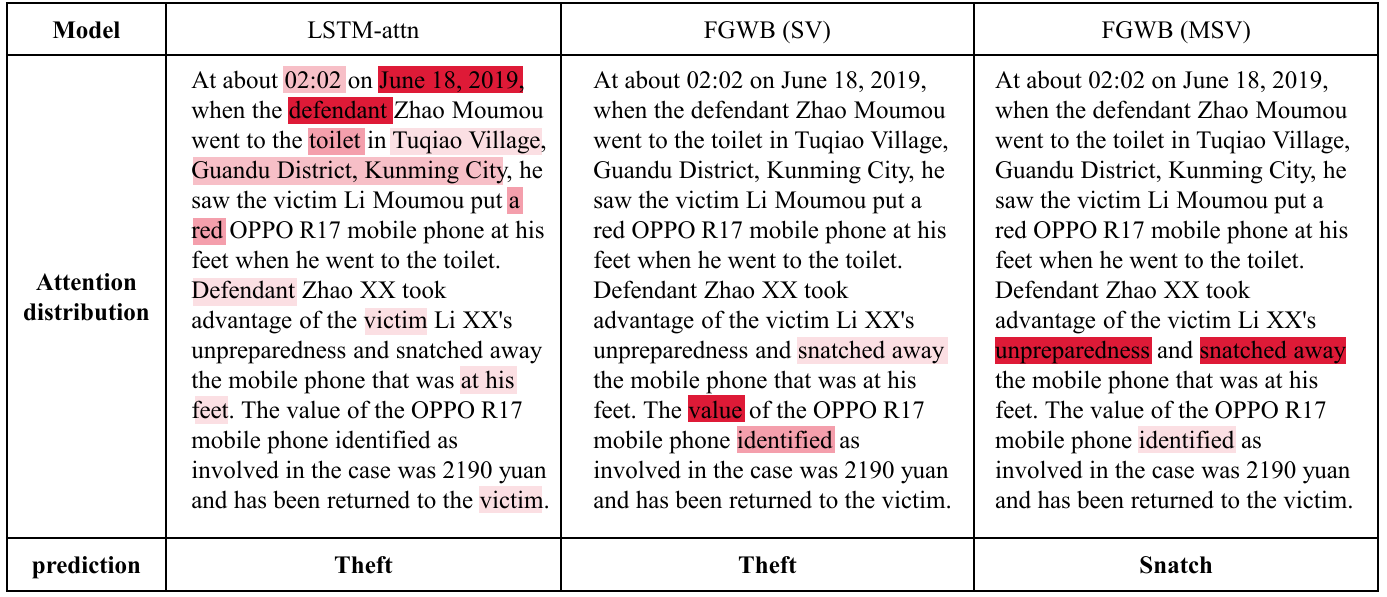}
\caption{Attention distribution from different models for a \emph{Snatch} case. ``SV" stands for using attention supervision, while ``MSV" stands for using multi-attention supervision. They are both implemented on LSTM.}
\label{attention}
\end{figure*}

\paragraph{Result of Ablation Experiment:}
From Tab. \ref{ablation}, We conclude that: (1) ``w/o SV" results in a significant decrease in performance for both implementation methods compared to FGWB suggesting that attention supervision is effective in enhancing the model's ability to distinguish between confusing charges. Additionally, the decrease is more pronounced for LSTM, which is due to the inferior performance of LSTM compared to Electra. This implies that attention supervision has a more substantial impact on improving LSTM's performance. (2) When it comes to ``w/o Multi-Attn", there is a significant performance decrease compared to FWGB. This implies that the multi-attention mechanism successfully provides separate attention spaces for each criminal charge, avoiding mutual interference and achieving better supervision results. (3) When it comes to ``w/o KG", there is a performance decrease compared to FWGB. This highlights the significance of knowledge graph assistance in constructing the word bag. The knowledge graph, summarized by legal experts, retains crucial elements that can differentiate between criminal charges. Filtering keywords from the knowledge graph's components is, in fact, an effective form of external knowledge incorporation, aiding the model in learning expert knowledge to distinguish between confusing charges.

\paragraph{Performance by the Coefficient for Attention Supervision $\lambda$:}
\begin{figure}[t]
\centering
\includegraphics[width=0.5\textwidth]{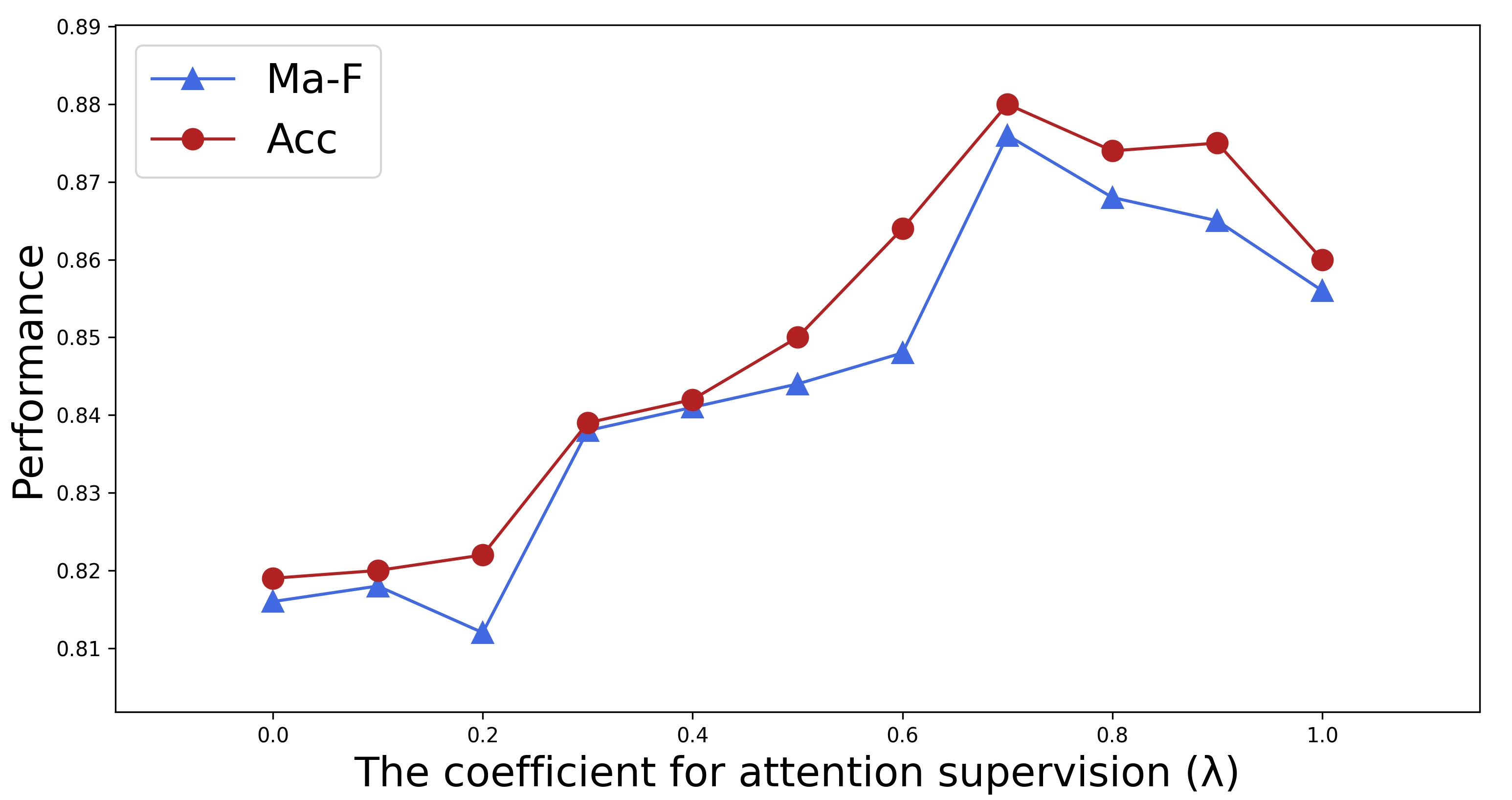}
\caption{Model performance by the coefficient $\lambda$ for attention supervision.}
\label{coefficient}
\end{figure}
We investigate the impact of changing the coefficient $\lambda$, which controls the attention supervision loss, on the model FGWB (LSTM) in terms of Ma-F and Acc metrics. As shown in Fig. \ref{coefficient}, as $\lambda$ gradually increases, the model's performance exhibits an initial improvement followed by a decline, reaching its optimal performance at $\lambda=0.7$. This outcome indicates that attention supervision is effective in confusing charge prediction.

\subsection{Case Study}
Fig. \ref{attention} shows the heat maps of a real \emph{Snatch} case when predicting the charges by the LSTM-attn model, FGWB (SV), and FGWB (MSV) respectively. Words with a deeper background color have higher attention weights. We observe that: (1) In the LSTM-attn model, we observe that it pays attention to many irrelevant details, such as the crime time ('June 18, 2019'; '02:02'). Due to the model's attention not being focused on crucial information, it made a prediction error, classifying the crime as \emph{Theft}.
(2) FGWB (LSTM+SV) identifies keywords related to the charge like \emph{value}, \emph{identified}, and \emph{snatch away}. However, it assigns incorrect weights to these words, leading to a false theft charge prediction.  This occurs because when all charges are supervised with a shared attention mechanism, they tend to influence each other, emphasizing information they have in common while potentially neglecting differentiating details.
(3) FGWB (MSV) places emphasis on \emph{unpreparedness} and increases the attention weight on \emph{snatch away}, both of which are constituent elements of \emph{Snatch} associated with \emph{violence against people}. As a result, FGWB, following the logic akin to that of a judge, correctly predicts \emph{Snatch}.

\section{Conclusion}
In this paper, we address the challenging task of confusing charge prediction within the legal domain. Existing charge prediction methods often fall short of effectively distinguishing between easily confused charges. Our innovative approach, the "From Graph to Word Bag (FWGB)" model, leverages constituent elements within a legal knowledge graph to enhance predictive accuracy. We introduce a multi-attention supervision mechanism to ensure that the model focuses on critical information within the context, leading to substantial improvements in performance. Through extensive experimentation, we have validated the effectiveness of our approach using real-world judicial documents.

\section{Ethical Discussion}
\label{app:discussion}
Automatic charge prediction is a sensitive field of AI. While our goal is to surpass the performance of existing approaches, it's essential to acknowledge that these technologies are not yet ready for practical implementation. Legal cases often contain sensitive personal information, highlighting the importance of protecting privacy when processing datasets \cite{XU202366}. Ensuring the ethical deployment of artificial intelligence systems in legal decision-making requires strict safeguards, transparency, and sustained ethical considerations to protect individual rights and maintain trust in the legal system. Additionally, exploring more suitable encoding methods to mitigate biases introduced by data distributions can promote fairness \cite{community_preserve}.

\section{Acknowledgements}
This work was supported in part by National Key Research and Development Program of China (2022YFC3340900), National Natural Science Foundation of China (No. 62376243, 62037001, U20A20387), the StarryNight Science Fund of Zhejiang University Shanghai Institute for Advanced Study (SN-ZJU-SIAS-0010) and Project by Shanghai AI Laboratory (P22KS00111).

\nocite{*}
\section{Bibliographical References}\label{sec:reference}

\bibliographystyle{lrec-coling2024-natbib}
\bibliography{lrec-coling2024-example}

\section{Language Resource References}
\label{lr:ref}
\bibliographystylelanguageresource{lrec-coling2024-natbib}
\bibliographylanguageresource{languageresource}
\end{document}